\def\year{2022}\relax
\documentclass[letterpaper]{article} % DO NOT CHANGE THIS
\usepackage{aaai22}  % DO NOT CHANGE THIS
\usepackage{times}  % DO NOT CHANGE THIS
\usepackage{helvet}  % DO NOT CHANGE THIS
\usepackage{courier}  % DO NOT CHANGE THIS
\usepackage[hyphens]{url}  % DO NOT CHANGE THIS
\usepackage{graphicx} % DO NOT CHANGE THIS
\usepackage{natbib}  % DO NOT CHANGE THIS AND DO NOT ADD ANY OPTIONS TO IT
\usepackage{caption} % DO NOT CHANGE THIS AND DO NOT ADD ANY OPTIONS TO IT
\DeclareCaptionStyle{ruled}{labelfont=normalfont,labelsep=colon,strut=off} % DO NOT CHANGE THIS
\usepackage{algorithm}
\usepackage{algorithmic}

\usepackage{times}
\usepackage{epsfig}
\usepackage{graphicx}
\usepackage{amsmath}
\usepackage{amssymb}
\usepackage{array}
\usepackage[hidelinks]{hyperref}

\usepackage{newfloat}
\usepackage{listings}
\floatstyle{ruled}
\newfloat{listing}{tb}{lst}{}
\floatname{listing}{Listing}
\title{Towards Autonomous Satellite Communications: An AI-based Framework to Address System-level Challenges}
\author{
    Juan Jose Garau-Luis\equalcontrib, Skylar Eiskowitz\equalcontrib, Nils Pachler\equalcontrib,\\
    Edward Crawley, Bruce Cameron
}
\affiliations{
    Engineering Systems Laboratory \\
    Massachusetts Institute of Technology\\
    \{garau, eiskowit, pachler, crawley, bcameron\}@mit.edu
}

\begin{document}

\maketitle

\begin{abstract}
The next generation of satellite constellations is designed to better address the future needs of our connected society: highly-variable data demand, mobile connectivity, and reaching more under-served regions. Artificial Intelligence (AI) and learning-based methods are expected to become key players in the industry, given the poor scalability and slow reaction time of current resource allocation mechanisms. While AI frameworks have been validated for isolated communication tasks or subproblems, there is still not a clear path to achieve fully-autonomous satellite systems. Part of this issue results from the focus on subproblems when designing models, instead of the necessary system-level perspective. In this paper we try to bridge this gap by characterizing the system-level needs that must be met to increase satellite autonomy, and introduce three AI-based components (Demand Estimator, Offline Planner, and Real Time Engine) that jointly address them. We first do a broad literature review on the different subproblems and identify the missing links to the system-level goals. In response to these gaps, we outline the three necessary components and highlight their interactions. We also discuss how current models can be incorporated into the framework and possible directions of future work.
\end{abstract}

\section{Introduction}

The satellite communications landscape has been undergoing significant changes during the last few years. The growth of streaming platforms and other data-intensive services has pushed a television/video-centered industry into a data-dominated market characterized by higher throughput demands \cite{NorthernSkyResearch2019}. In addition, new ground stations in unserved communities where terrestrial networks are not an affordable option \cite{Reut}, as well as new mobile terminals in almost every plane, make user bases larger and more complex. In response, highly-flexible megaconstellations are being deployed, which will flood the market in the upcoming years. What once was a static process from a resource control point of view is becoming increasingly dynamic and more challenging.

\begin{figure*}[t]
\begin{center}
\includegraphics[width=0.95\linewidth]{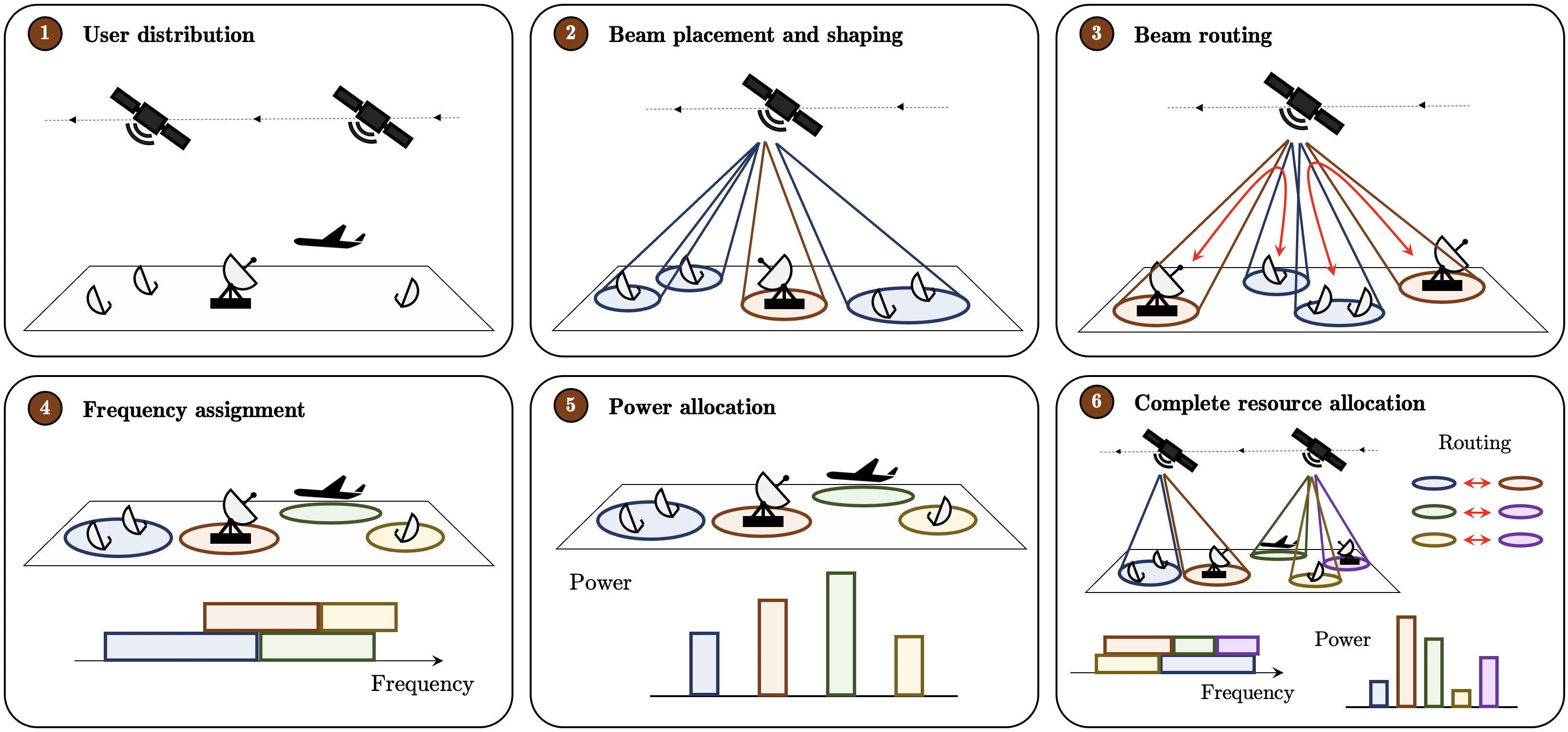}
\end{center}
\caption{Resource allocation problem in satellite communications divided into a sequence of subproblems. From a set of fixed and mobile users with certain demand requirements (step 1), operators first decide how many beams to use, as well as their location and shape (step 2). Then, each beam is routed to a gateway (step 3) and a certain amount of bandwidth within the available frequency spectrum is allocated (step 4). The final subproblem involves powering each beam (step 5), resulting in a complete resource allocation (step 6).}
\label{fig:ProblemDiagram}
\end{figure*}

Satellite operators have been able to incorporate the necessary hardware improvements to adapt to this new dynamic context. Beamforming capabilities capture more complex terrestrial landscapes, digital payloads allow for real time power and bandwidth control \cite{Balty2009, Angeletti2008}, and the industry is entering an era marked by mega constellations that add new degrees of freedom \cite{Vidal2021}. However, current resource management policies are human-driven, which is no longer sufficient given the high-dimensionality and flexibility of upcoming systems. Therefore, operators are looking to outsource many of these human decisions to autonomous agents, as it will be the key to being competitive in this growing market \cite{Coleman2019}.
% ; more complex control processes must be developed to manage these new capabilities, especially

Artificial Intelligence (AI) and learning-based solutions are prominent in satellite communications research as a promising way to control these complex space systems. From finding optimal policies to steer antennas \cite{Abbas2015}, to leveraging time series data to identify contingency scenarios \cite{Ackerman2020}, current research suggests that satellite constellations can achieve increasing levels of automation.

While AI is capable of matching the state-of-the-art performance in many of these specific tasks \cite{Luis2020, Aravanis2015, Paris2019, Lei2010, Durand2017, Ferreira2018, Luis2019}, there is not a clear path to move the focus from solving specific tasks to the design of autonomous satellites able to achieve system-level goals. Part of the hardware components studied in recent papers such as antennas or power amplifiers are not isolated, but are part of larger entities that interact, as thoroughly studied by the space systems community \cite{Angeletti2008}. In that sense, there is little consideration on how the proposed AI models will fit into the whole constellation architecture.

Although scarce, some papers do attempt to bridge that gap. This is the case of \cite{Guerster2019}, where a Dynamic Resource Management (DRM) system concept is presented to specifically address the adapted process to control different resources such as power and bandwidth on a satellite. Authors divide their system into a deployed physical network and a short-term and long-term planners that make resource allocation decisions based on demand predictions. While the overall goal of this paper is well aligned with the industry needs, there is little emphasis on how the AI and learning-based technologies on a subsystem level connect to their role on the system level and vice versa.

In this paper we try to close this research gap by proposing a similar resource management framework to the one presented in \cite{Guerster2019}, highlighting the necessary components in an architecture aimed at solving the complete DRM problem in satellite communications. We identify three interconnected subsystems, namely the Demand Estimator, Offline Planner, and Real Time Engine, that jointly affect the performance of the whole system on different resource management tasks such as power allocation, beam routing, and frequency assignment. Our goal is to analyze the potential value of different AI algorithms on both the system and subsystem levels and understand their tradeoffs. Finally, we outline prevailing challenges to achieve better autonomous constellations and areas of future work for each task.

\section{Dynamic Resource Management in Multibeam Satellite Constellations}
\label{sec:2}

The DRM problem in a satellite communications constellation consists of finding a feasible distribution of the constellation's resources that satisfies the changing demand needs of a user network. Fully solving this problem entails addressing a sequence of complex subproblems, each related to one of the resources to allocate, as illustrated in Figure \ref{fig:ProblemDiagram}. Overall, it is a time-sensitive problem, as the dynamic behavior of the users' demand requires updating parts of the solution to the subproblems in real time, as pictured in Figure \ref{fig:DiagramUpdate}. Given these time constraints and the high-dimensionality context of upcoming constellations, AI and learning-based algorithms are seen as essential to keep up with the constantly-changing demand in a way that won't require operators to severely over-provision resources. 
% Talk 1 paragraph (I'd say 4-5 sentences) about the problem's scope (What do we need to do?). The idea is to mention the two agents (users and resources), so that we can explain them in the following lines.

\subsection{User terminals}
A user (represented by a small antenna or a plane in Figure \ref{fig:ProblemDiagram}) is defined as any entity connected to the constellation that expects certain throughput demand needs to be met (e.g., a person watching a movie through an online streaming platform). Two factors characterize each user: demand and position. In this context, we assume that the user's demand is flexible and may vary over time depending on the user's behaviour. For example, users navigating the web tend to generate highly-fluctuating demand, while backhauling customers tend to be more steady. The position of a user can either be fixed (e.g., common households antennas) or change over time (e.g., planes and ships).

% Talk about the users (types of services, mobility vs non-mobility, demand over time, etc).

\subsection{Constellation's resources}
To satisfy the users' demand, operators need to resolve four subproblems in order to distribute the constellation's resources: beam placement and shaping, gateway routing, frequency assignment, and power allocation.

\begin{figure}[t]
\begin{center}
\includegraphics[width=0.99\linewidth]{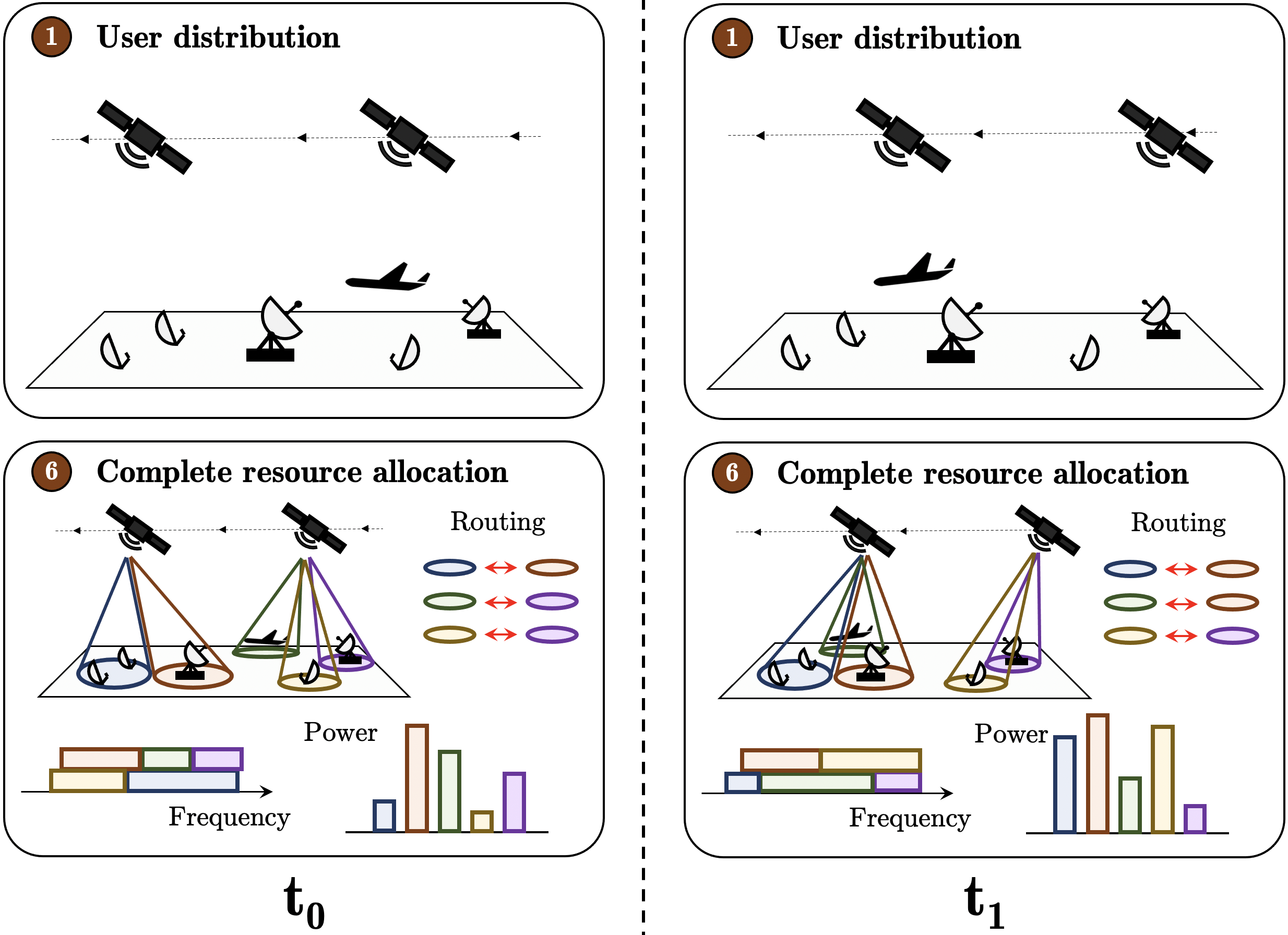}
\end{center}
\caption{User distributions and resource allocations changing over time. A change in the user distribution from time $t_0$ to time $t_1$ prompts a change in the allocation of the constellation's resources.}
\label{fig:DiagramUpdate}
\end{figure}

\textbf{Beam placement and beam shaping.}
Shown in the second step in Figure \ref{fig:ProblemDiagram}, the beam placement problem consists of achieving a feasible mapping of users to satellite beams. A beam is a communication channel between one or more users and a satellite. Different users can be assigned to the same beam only if they are all on the beam's terrestrial footprint, which is influenced by the beam's shape. Therefore, this process entails grouping users together. Also represented in the second step in Figure \ref{fig:ProblemDiagram}, the beam shaping problem consists of deciding the shapes and sizes of the beams, and therefore their footprints. Depending on the context, operators might prefer to use few beams with little overlap (which reduces interference) or a higher number of beams with simpler shapes (which is more energy-efficient). The beam shaping problem is not independent from the beam placement problem, so they need to be solved simultaneously. 

\textbf{Gateway routing.}
Depicted in the third step in Figure \ref{fig:ProblemDiagram}, the gateway routing problem consists of mapping beams to gateways. A gateway, shown as a bigger antenna in the figure, is an operator's ground station that gathers the traffic from many beams to the Internet and vice versa. There are multiple gateways connected to the network. How beams are mapped to gateways might be constrained by users' characteristics or preferences, and has a potential effect on the quality of the service.

\textbf{Frequency assignment.} 
Represented by the fourth step in Figure \ref{fig:ProblemDiagram}, the frequency assignment problem consists of assigning a central frequency and an amount of bandwidth to each beam. This goes beyond simply partitioning the frequency spectrum, since two beams can use overlapping frequencies due to several frequency reuse mechanisms available in each satellite. The extent to which two beams can use overlapping frequencies depends on how close their terrestrial footprints are.

\textbf{Power allocation.}
Finally, the fifth step of the process shown in Figure \ref{fig:ProblemDiagram} is power allocation, which sets the transmitted radio-frequency (RF) power to every beam. Generally, satellites are power-constrained, but there can be additional restrictions that limit the amount of power available to each beam. Operators can then reduce the necessary power of a specific beam by increasing its bandwidth, for instance.

\subsection{System-level challenges}
\label{sec:2.3}
From a system's perspective, the emerging challenges when considering the complete DRM problem are threefold: 
\begin{itemize}
\item First, each of the presented subproblems is not isolated; the decisions made at a certain step of the process impact the steps to come. For example, placing fewer beams (step 2 in Figure \ref{fig:ProblemDiagram}) might make routing easier (step 3) but requires higher power consumption (step 5).
\item Another challenge comes from the interactions between different satellites in the constellation. One example is the handover process, where a beam serving a user switches from satellite to satellite due to constellation and user dynamics (as shown in Figure \ref{fig:DiagramUpdate} for the plane). 
\item Finally, given the dynamic nature of the environment (e.g., changes in user demand patterns, moving beams, new users, or system failures), operators constantly reevaluate resource management decisions in real time. This is shown in Figure \ref{fig:DiagramUpdate}, where a change in the user distribution from time $t_0$ to $t_1$ requires a reconsideration of the resource allocation decisions. From an algorithmic perspective it is challenging to recompute an allocation given these time constraints and the high-dimensionality of the constellations.
\end{itemize}

In response to these challenges, AI and learning-based algorithms have the potential to play a substantial role in solving the resource allocation problem. This has started to be explored mainly at the subproblem level, mostly with a focus on brute performance. In the following section we discuss the specific AI models that have been proposed to address each subproblem.

% TABLE TEMPLATE
\begin{table*}
\begin{center}
\begin{tabular}{|l|l|m{3cm}|m{2cm}|m{2cm}|m{2cm}|}
\hline
\multicolumn{1}{|c|}{\textbf{Method}} & \multicolumn{1}{c|}{\textbf{Algorithm}} & \textbf{Beam plac. \& shape}  & \textbf{Routing}   & \textbf{Freq. assign.}   & \textbf{Power alloc.}   \\ \hline
  & \textbf{Genetic Algorithm}    & \citeyear{wangdynamic, Deng2018, He2017, Anzalchi2010, Angeletti2006} &  \citeyear{rao2011performance}  & \citeyear{He2017, tirmizi2015efficient, Angeletti2006, Paris2019} & \citeyear{Aravanis2015, Anzalchi2010, Paris2019} \\ \cline{2-6} 
\textbf{Metaheuristics}  & \textbf{Particle Swarm Opt.}   &   &     & \citeyear{pachler20b}  & \citeyear{Durand2017, pachler20b} \\ \cline{2-6} 
 & \textbf{Simulated Annealing} & \citeyear{Camino2014} &  & \citeyear{vidal2020joint, Cocco2018, Camino2014} & \citeyear{vidal2020joint, Cocco2018}   \\ \hline
\textbf{Reinforcement} & \textbf{Q-learning}  &  \citeyear{hu2020dynamic}  &  \citeyear{gong2020adaptive}  & \citeyear{liao2020distributed, zheng2020leo, Ferreira2018, Hu2018ASystems}   & \citeyear{zhang2020online, liao2020distributed, Ferreira2018}  \\ \cline{2-6} 
\textbf{Learning}   & \textbf{Policy Gradient}  &   &    &    & \citeyear{Luis2019, Luis2020} \\ \hline
\textbf{Supervised learn.}  & \textbf{Neural architectures}  &  &  & \citeyear{Funabiki1997, Salcedo-Sanz2005}  & \\ \hline
\end{tabular}
\end{center}
\caption{AI methods and algorithms applied to each resource allocation subproblem in the literature. Studies might focus on the complete subproblem or solve it partially. Some studies address parts of two different subproblems.}
\label{tab:literature}
\end{table*}

\section{Related Work}

Multiple studies have addressed solving specific resource allocation subproblems by means of AI and learning-based methods. Most of the literature focuses on metaheuristic algorithms given their straightforward frameworks, but in the recent years authors have started to rely more on neural networks and Reinforcement Learning agents. A summary of the literature can be found in Table \ref{tab:literature}; we now discuss each subproblem separately.

\subsection{AI and learning-based methods for resource allocation subproblems}

\textbf{Beam placement and beam shaping.} Most studies that approach the beam placement problem in the literature focus on beam scheduling, i.e., defining start and end times for each beam \cite{wangdynamic, Deng2018, He2017, Anzalchi2010, Angeletti2006}. All these works show efficient Genetic Algorithms (GA) that optimize based on user satisfaction, total capacity, and/or power usage in the context of satellites with tens of beams. Some authors \cite{hu2020dynamic} have also proposed Reinforcement Learning (RL) techniques, specifically Q-learning algorithms, to deal with the added complexity when moving to the thousand-beam range. Regarding other beam placement-specific tasks, researchers have recommended Simulated Annealing (SA) \cite{Camino2014} to specifically solve the question of how many beams to place and where. Although the beam shaping subproblem has been addressed by means of heuristics and mathematical programming frameworks \cite{kyrgiazos2013irregular, camino2016mixed}, we could not find AI-based proposals for this task.

% The main difference between these works is the objective function. Specifically, Wang \cite{wangdynamic} proposes a multi-objective formulation to maximize system throughput while minimizing packet delay. Deng \cite{Deng2018} suggests using maximum number of tasks, minimum number of changes in the scheduling scheme and minimum number of sub-divisions in the tasks as figures of merit. He \cite{He2017} contrasts the resources allocated to voice and non-voice services. On a full-system design, Anzalchi \cite{Anzalchi2010} tries to minimize both the unmet system capacity (USC, as a measure for user satisfaction) and total power. Finally, Angeletti \cite{Angeletti2006} proposes a formulation to maximize capacity while ensuring constraint satisfaction. In a similar line \cite{hu2020dynamic} suggests a Q-learning Deep Reinforcement Learning (DRL) model to optimize packet delay, maximum non-voice throughput, and total fairness at the same time. Center lat/lon and number of beams with SA \cite{Camino2014}. [The tasks can be center lat/lon, beam-task scheduling, number of beams, and beam shape]

\textbf{Routing.} The scarce literature on beam and gateway routing mainly focuses on how to route users to gateways in order to minimize delay. While most studies suggest non-AI load balancing solutions \cite{torkzaban2020joint, yang2016towards}, some authors have proposed GA \cite{rao2011performance} and Q-Learning \cite{gong2020adaptive} frameworks in the context of highly-varying environments.

\textbf{Frequency assignment.} From the four subproblems that constitute the complete resource allocation process, the frequency assignment is the most studied in both depth and breadth. In general terms, when all beams use the same bandwidth, this problem can be understood as assigning the central frequency for each beam. Both metaheuristic (GA, \cite{He2017, tirmizi2015efficient}, SA, \cite{vidal2020joint}) and RL-based (Q-learning, \cite{Hu2018ASystems}) formulations have proven successful in solving this task. Some studies do consider flexible bandwidth and address it with both metaheuristic frameworks (GA, \cite{Angeletti2006, Paris2019}, SA \cite{Cocco2018}) and learning-based models (Q-Learning, \cite{liao2020distributed, zheng2020leo}, supervised learning with neural networks, \cite{Funabiki1997, Salcedo-Sanz2005}). While metaheuristics are better suited for scenarios with fewer beams, the high-dimensionality of real operations makes learning-based techniques a competitive solution for time-constrained contexts. Specific instances of the independent bandwidth allocation problem (i.e., choosing bandwidth fractions) have also been studied through an AI lens (PSO, \cite{pachler20b}, Q-Learning, \cite{Ferreira2018}). Finally, although most models consider a single pool of frequency, some research has focused on the effects of frequency reuse and multiple polarizations (SA, \cite{Camino2014}).

\textbf{Power allocation.} Finally, a large number of studies address the power allocation problem by means of AI and learning-based methods, where the goal is to autonomously choose how much power to provide to each beam while respecting beam- and satellite-wise maximum power constraints. Metaheuristic algorithms (GA \cite{Aravanis2015, Anzalchi2010, Paris2019}, PSO \cite{Durand2017, pachler20b}, and SA \cite{vidal2020joint, Cocco2018}) prove to be effective ways to meet these constraints. However, it is hard for them to scale up to thousands of beams and still converge in the range of seconds. Consequently, authors have also looked into RL-based methods, specifically Q-Learning \cite{zhang2020online, liao2020distributed, Ferreira2018} and Policy Gradient \cite{Luis2019, Luis2020}.

As seen in Table \ref{tab:literature}, some models address decisions concerning more than one subproblem. For example, the Genetic Algorithm proposed in \cite{Angeletti2006} encompasses both beam placement and frequency assignment decisions -- but it does not completely address either of them. In most studies, specific subproblems or use cases have been simplified in order to achieve convergence within reasonable time. It remains unclear the best way in which AI and learning-based techniques can expand to encompass multiple subproblems in high-dimensional and time-restricted scenarios.

\subsection{System-level gaps in current approaches}

%Paragraph 1: there are gaps in the literatuer, subsystem entabglement bullet 1
Despite the progress on the specific subproblems, gaps in the literature are evident at the system level, where challenges outlined in Section \ref{sec:2.3} remain untouched. First, studies often analyze isolated subproblems, but do not consider the dependence on the next subproblem. For example, authors in \cite{Camino2014} need to rely on a demand-related metric instead of a user satisfaction metric because their model does not consider power allocation (last step of the resource allocation process). It remains unclear how both metrics relate without solving the latter subproblem. As discussed in the previous section, when more than one subproblem is addressed, most of the time those are not completely solved, and simplifications are applied. 

%Paragraph 2: second bullet point
Next, studies fail to address the satellite as a part of a larger constellation network, and so they avoid the added complexity of this higher-dimensional problem. For example, most frequency assignment algorithms (e.g., \cite{vidal2020joint, liao2020distributed}) overlook that, given the constellation dynamics, frequency must be assigned to beams that are constantly changing. An allocation that is valid for a specific point in time may have severe limitations at subsequent times, as different beams are handed over to a satellite following routing decisions. 

%Paragraph 3: an architecture needs to look at all subsystems
Finally, any DRM problem framework must also situate the subproblems and their interactions in the context of realistic operating scenarios.  Most studies assume a given future user distribution \cite{Aravanis2015, Cocco2018}, which fails to capture the reality that in many cases users' future demand and location are unknown. Forecasting user demand for a short upcoming horizon leads to less uncertainty, but also less time to react and utilize this new knowledge. Challenges like this are overlooked when studies don't simulate the user distribution of the problem. Furthermore, studies omit the increasing uncertainty that directly comes from having higher dimensionality, leading to optimistic results and solutions that might actually be infeasible due to the uncaptured uncertainty.

%\section{A title like here is the architecture we consider}
\section{Addressing system-level complexity with an AI-based framework}
\label{sec:4}
% Potential section contents:
% \begin{itemize}
%     \item System overview, why this consideration makes sense
%     \item The demand estimator: what is it and why it needs to rely on learning
%     \item The capacity planner optimizer: what is it and why it needs to rely on AI (here I would consider AI as in the Russell book, i.e., more than just ML and trees)
%     \item The real-time engine: what is it and why it needs to rely on learning
%     \item Watch out these systems interact: Why we need to consider a systems perspective, the importance of designing algorithms from this perspective
% \end{itemize}

\begin{figure*}[t]
\begin{center}
\includegraphics[width=0.8\linewidth]{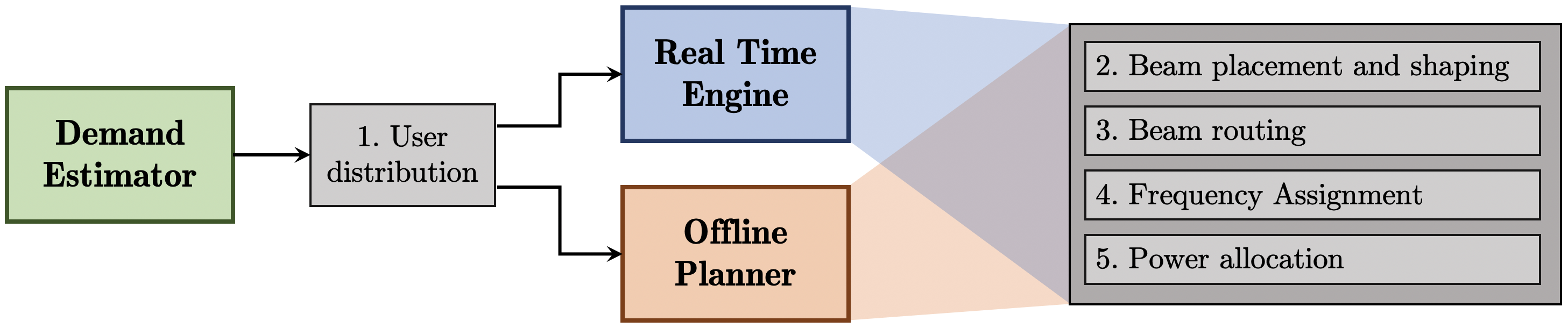}
\end{center}
\caption{Necessary components in a resource management system. The \textit{Demand Estimator} is responsible for predicting changes in the user distribution (demand and location) at different time horizons. The \textit{Offline Planner} then addresses long-term resource allocation tasks, while the \textit{Real Time Engine} works on a time scale of seconds to minutes, fixing any service failure and leveraging resource-efficient opportunities.}
\label{fig:BlockDiagram}
\end{figure*}

In this section we introduce the necessary elements of a resource management system and explore the role of AI and learning-based methods to address the system-level complexities. The proposed elements are similar to the ones introduced in \cite{Guerster2019}, and correspond to three components: the Demand Estimator (DE), responsible for forecasting the user distribution; the Offline Planner (OP), responsible for long-term proactive decisions; and the Real Time Engine (RTE), responsible for short-term reactive decisions. These three subsystems, as depicted in Figure \ref{fig:BlockDiagram}, jointly address the four resource allocation subproblems. We also discuss how these systems should interact to meet all the constraints imposed by real-time operation and large constellations.

\subsection{System goals}

%remind reader what system challenges are that have not been solved by literature
As described in Section \ref{sec:2.3}, three challenges remain unaddressed from a high-level problem perspective: 1) subproblems need to be considered simultaneously, 2) satellites need to be considered integrated in a larger entity, the constellation, and 3) the environment is dynamic and high-dimensional.

%focus on the third one because the first 2 are obvious (systems interactions)
%now we have abstracted what this third problem means for waht an archiecture must be able to handleLuncertainty and limited compuational time
From a real operation point of view, the last point requires operators to rely on user distribution forecasts instead of perfect information to decide on the resource allocation. Operators can decide to predict future distributions with a specific level of uncertainty depending on how far the prediction horizon is. While a lower uncertainty given by shorter horizons is best for finding better allocations, it can substantially reduce the time windows available to compute them. These two features (uncertainty and computing time) define the challenges a DRM framework must operate under. Any architecture for this resource allocation system must be able to address different levels of uncertainty and computing windows.

\subsection{DRM Framework overview}

Our framework, depicted in Figure \ref{fig:BlockDiagram}, consists of three components that jointly aim to solve the complete resource allocation problem described in Figure \ref{fig:ProblemDiagram} and address the necessary real-time changes. This process starts with the Demand Estimator forecasting the user distribution (which includes the demand and location of users) at future horizons. The long-term predictions are passed to the Offline Planner, and the short-time predictions are sent to the Real Time Engine.

In conjunction, both the OP and RTE are responsible for the four subproblems. While they both can act on the same set of decisions, the objective and responsibilities of each subsystem are significantly different. The goal of the OP is to compute complete allocations from scratch ahead of time based on user distribution estimates. Given the problem's high-dimensionality and complexity, the time needed to run this process is not negligible, which limits the OP to long-term decisions where computing time is not a constraint. However, since the long-term demand predictions are uncertain, the real and estimated demand may significantly differ. The objective of the RTE is then to tune the resource allocation originally computed by the OP to resolve these discrepancies based on more detailed knowledge of the short-term user distribution. To that end, the RTE should first decide which specific resources it will change, as there might be more than one possible alternative. The RTE iteratively repeats that process until the next OP update.

\subsection{AI-based decision-making in the subsystems}

\textbf{Demand Estimator.} The DE is tasked with producing forecasts that predict changes in user distributions as accurately as possible, so that the OP and RTE avoid under- and over-provisioning resources. Due to the increasingly complex user demand patterns, it may no longer be sufficient to rely on simple statistical methods that are unable to model complex, nonlinear behaviours \cite{Suganthi2012}. Following similar trends in other industries \cite{Fildes2002, Zhao2012, Wang2017, Panapakidis2016, Sagheer2019} time series forecasting models like recurrent neural networks and attention-based mechanisms \cite{Vaswani2017} are essential to accurately predict these patterns. These methods take advantage of large amounts of data to constantly improve their performance and capture both short- and long-term patterns \cite{Vaswani2017}. Furthermore, novel meta learning algorithms may be able to leverage little data coming from new and out-of-distribution users \cite{Panigrahi2021}.

\textbf{Offline Planner.} The OP is responsible for creating a feasible long-term resource distribution plan that satisfies the forecasted users' demand. Given that this problem is high-dimensional and complex, conventional resource allocation techniques such as Lagrangian-based binary search for frequency assignment \cite{park2012dynamic} or heuristic-based power allocation \cite{destounis2011dynamic} are unlikely to succeed, even on a subproblem level. In this line, metaheuristics have proven successful in solving each of the individual subproblems \cite{Deng2018, rao2011performance, tirmizi2015efficient, Durand2017}, and show promising results when dealing with multiple subproblems \cite{Camino2014, pachler20b, He2017}. In cases where the dimensionality is so high that these techniques are no longer practical, learning-based alternatives have been explored \cite{hu2020dynamic, gong2020adaptive, wangdynamic, Luis2019}.

\textbf{Real-Time Engine} The RTE has to refine the resource allocation proposed by the OP based on the discrepancies between the real and forecasted user distributions. The goal of the RTE is not to allocate on all possible variables, but rather to a lower-dimensional subset that does not compromise a fast convergence time. In that sense, the Reinforcement Learning-based agents presented in the literature could be better suited to reallocate power \cite{zhang2020online, liao2020distributed, Ferreira2018, Luis2019, Luis2020}, reassign frequencies \cite{liao2020distributed, zheng2020leo, Ferreira2018, Hu2018ASystems}, reroute \cite{gong2020adaptive}, or rearrange the beam placement \cite{hu2020dynamic}. In the case of routing and beam placement tasks, the lack of more studies also opens the door to consider metaheuristic-based alternatives, such as \cite{wangdynamic, Deng2018, He2017, Anzalchi2010, Angeletti2006, rao2011performance}.

\subsection{System interactions} \label{sec:4.4}
%what are scenarios that make task harder for AI
 
% First paragraph: There is interaction, specifically around the pairs, we summarize it below.
The proposed DRM framework hinges on the distribution of functionalities across the three subsystems, and, as a result, it is important to study how the DE, OP, and RTE interact. Here, we discuss these interactions across the pairs and how AI can handle them.

\textbf{Demand Estimator and OP/RTE}. The DE interacts with both the OP and RTE by feeding in forecasts of the future user distribution. The key challenge the DE must deal with is uncertainty. To quantify uncertainty, common metrics (e.g., MAPE, MAE, and MSE) measure how well a model behaves on a test dataset, but more important to our application are measures of under- and over-predictions in demand. Satellite operators typically overprovision, because an underestimation of future user demand is unacceptable in serving their customers. The DE itself may account for this preference by training a model with an asymmetric loss function that places a heavier penalty on underestimations. Furthermore, the forecasts being probabilistic instead of point estimates gives insight into the amount of uncertainty at the prediction level. This allows the OP to better plan for worst-case scenarios and reduce the number of times the RTE needs to take action. To avoid producing overly-conservative allocations, the OP and RTE subsystems could also incorporate robustness mechanisms against uncertainty, as explored in the literature for both metaheuristic frameworks \cite{keane1995genetic, al2006incorporating} and RL \cite{bellemare2017distributional, dabney2018implicit}.

%What happens with overestimation and underestimation. What is the effect of overestimation and underestimation on OP/RTE. Possible ways to mitigate that effect. Citations on uncertainty in metaheuristics and uncertainty in RL. 

\textbf{OP and RTE}. The interaction between OP and RTE is more subtle since it is a consequence of the shared control of the constellation's resources. While the OP does a long-term recurrent allocation of all resources, the RTE should only react when nominal demand predictions are out of distribution. We envision this happening under six circumstances: 1) demand spike, when the real demand is significantly higher than expected; 2) demand diminish, when real demand is significantly lower than expected; 3) new user, when a new user connects to the system using a new (non-allocated) beam; 4) mobility re-route, when a ship or plane follows a different route than the one expected; 5) gateway failure, when a gateway antenna temporarily or permanently fails; and 6) satellite failure, when a satellite temporarily or permanently fails. Both the RTE's type of action and time to react significantly vary depending on the event. Nevertheless, all these situations can be considered as out-of-distribution occurrences, which have traditionally been a challenge to AI and learning-based models as a form of non-stationarity or high stochasticity. Recent studies \cite{Peng2018, Mankowitz2019, Ghosh2017} propose generalization mechanisms which might be able to address these scenarios.

% Which are the events that trigger an RTE action, and which challenges do those pose for AI (e.g., demand spike is out-of-distribution). Worst case planning: Planning for CIR and the consequences: less new users, etc. 

The challenges emerging from system interactions pose additional problems to AI and have been traditionally under-studied in satellite communications literature. We consider these issues as one of the remaining open challenges when it comes to the Dynamic Resource Management problem and a necessary direction of future work. In the next section we extend on this idea and other open challenges and future work directions.

\section{Open Challenges}
%5.1 problem AI has on subproblem level: literature does this
%5.2 how AI must help in system level challenges
%5.3 what are next step towards automony: more flexibility (orbits as a resource)transfer learning

Different challenges remain regarding the use of AI and learning-based technologies for the Dynamic Resource Management problem. These can be analyzed from a subproblem perspective, a system-level perspective considering interactions, and from the perspective of autonomous systems. In this section we cover all three and propose directions of future work, highlighting those areas where AI is still not present.

% Introduction paragraph of what this section is about. (challenges in subproblems and interactions, as well as future work).

\subsection{Limitations on subproblem performance}
We first address the performance limitations that subproblem-oriented AI algorithms face and which recently-proposed frameworks could help mitigate them.
\begin{itemize}
    \item \textbf{High-dimensionality}. All four subproblems covered in this work (beam placement and shaping, gateway routing, frequency assignment, and power allocation) have been traditionally under-studied in high-dimensional scenarios \cite{Luis2020}, such as those dealing with tens of thousands of users or thousands of beams. Although some works do try to address this issue \cite{hu2020dynamic, Luis2019} by simply running tests on high-dimensional scenarios, recently-proposed mechanisms against high-dimensionality are still to be explored in the satellite communications context. This is the case, for instance, of specific RL frameworks \cite{van2020q, zahavy2018learn}.
    
    \item \textbf{Time limitations}. Time convergence is often an overlooked factor in current approaches, and can be especially critical for some of the metaheuristic-based algorithms presented. While most works focus on a static picture for which all resources can be allocated without a time horizon, future studies should incorporate new findings on speeding up the convergence time \cite{nia2009speeding}. Learning-based approaches are more robust against this issue given their quick online evaluations and offline training frameworks.
    
    \item \textbf{Constellation Dynamics}. Constraints imposed by constellation dynamics are usually neglected. Most works focus on single satellites, disregarding dependencies between the different satellites of the constellation and how they change over time. Constraint-based frameworks have been studied for both metaheuristic algorithms \cite{homaifar1994constrained} and RL \cite{garcia2015comprehensive, dalal2018safe, Bohez2019}.
    
    \item \textbf{Out-of-distribution events.} As highlighted in Section \ref{sec:4.4}, most events requiring the RTE involvement occur due to out-of-distribution events such as demand spikes or system failures. This issue has traditionally been a challenge for AI and learning-based methods, although recent works propose different methods to increase environment diversity \cite{Ghosh2017, Lee2019, Tobin2017} and consequently the agents' robustness.
\end{itemize}

\subsection{Limitations due to interactions}
When the DE, OP, and RTE act together, additional limitations might emerge. These are mainly related with forecasting quality and the OP-RTE interaction.
\begin{itemize}
    \item \textbf{Long-horizon forecasting}. The DE is faced with challenging prediction windows when a delay exists between the current and the target times. Uncertainty in these situations might be too large using traditional methods. Newer models that use attention mechanisms that learn complex patterns in data like \cite{Li2019} or \cite{Wu2020DeepCase} could be explored.
    
    \item \textbf{Multiuser prediction}. To forecast future user distributions and demands, the DE can do it on a per-user basis or rely on multivariate output models \cite{CharkrabortyKishanMehrotra1990ForecastingNetworks, Wan2019MultivariateForecasting}. The latter is not sufficiently-explored and can be the basis of frameworks to leverage low amounts of data (e.g., new users, non-stationary patterns) \cite{Yu2016TemporalPrediction}.
    
    \item \textbf{Search space complexity.} As a consequence of the OP-RTE interaction, once the OP sets a long-term resource allocation, the RTE will amend some parts of it as real-time operations take place. While maybe no more than tens of beams will be part of this process during a RTE cycle, the related search space will be notably more complex, as the RTE will be required to simultaneously make decisions on multiple variable types (e.g., beam center, bandwidth, power) instead of following a sequential process.  
    
\end{itemize}

% E.g., given sometimes RTE is given out-of-distribution data it needs to explore better ways to address that problem, it is especiallychallenging for RL, as seen in many numerous examples in the literature. 

\subsection{Areas of future work}
We finally address unexplored research directions that are related to more advanced problems in the context of autonomous constellations.
% orbit as a resource
% transfer learning

\begin{itemize}
    \item \textbf{Transfer learning for satellite architectures}. All models and algorithms so far have been designed in the context of specific satellite architectures. This assumption limits the application of one model to multiple satellite architectures. In the future, it is expected that satellites will have flexibility in their own configurations \cite{DeWeck2004OptimalAlgorithm}, to better address operators' needs. Being able to successfully transfer the models to these new configurations will be essential to maintain service quality.
    
    \item \textbf{New prediction models}. Historical data of current user bases might be rich enough to forecast using simple neural architectures. However, it is possible that this is not always the case, especially if new users choose on-demand services more frequently. In those cases, more complex neural architectures including Graph Neural Networks to leverage spatiotemporal data \cite{wang2020traffic}, attention mechanisms to improve performance \cite{Vaswani2017}, or transfer learning frameworks capable of few-shot-learning \cite{Panigrahi2021} might be better suited. 
    
    \item \textbf{Orbits as resources.} In this paper, like most studies on satellite communications, we assume all satellites are located on the same orbit. This might not always be the case, especially for megaconstellations. In that sense, we could consider allocating orbital resources as the fifth subproblem from a constellation perspective. It is likely that AI and learning-based methods can also be applied in that context.
    
    \item \textbf{Multiagent systems.} Finally, we want to highlight the possibility that the resource control is decentralized as opposed to a centralized DE, OP, and RTE. These scenarios align with the literature on multiagent systems \cite{foerster2018counterfactual, Rashid2018}, and satellite constellations are a specific use case that is starting to be explored \cite{he2020load, hu2020multi}.
\end{itemize}

\section{Conclusion}

In this work, we propose an AI-based framework to tackle the system-level challenges in the Dynamic Resource Management problem for satellite communications. This framework incorporates three necessary components (Demand Estimator, Offline Planner, and Real-Time Engine) to address the sequence of subproblems (beam placement and shaping, gateway routing, frequency assignment, and power allocation) that lead to the complete resource allocation. We identify potential component interactions that are often overlooked by current approaches and discuss why AI and learning-based methods are well suited to handle them. In this context, we examine the benefits of applying AI to solve the complete resource allocation problem, highlighting prevailing performance limitations that AI could overcome, as well as unexplored areas of future work.

\section*{Acknowledgments}
This work was supported by SES. The authors would like to thank SES for their input to this paper and their financial support. The project that gave rise to these results also received the support of a fellowship from ``la Caixa'' Foundation (ID 100010434). The fellowship code is LCF/BQ/AA19/11720036.

\bibliography{egbib}

\end{document}